\documentclass[letterpaper, 10 pt, conference]{ieeeconf}  

\IEEEoverridecommandlockouts                              

\usepackage{graphics} 
\usepackage{epsfig} 
\usepackage{mathptmx} 
\usepackage{times} 
\usepackage{amsmath} 
\usepackage{amssymb}  
\usepackage{tabularx} 
\usepackage{subfigure} 
\usepackage{booktabs}
\usepackage{tikz}
\usepackage[colorlinks=true]{hyperref} 

\def\checkmark{\tikz\fill[scale=0.4](0,.35) -- (.25,0) -- (1,.7) -- (.25,.15) -- cycle;} 

\title{\LARGE \bf
	Survey on LiDAR Perception in Adverse Weather Conditions
}

\author{Mariella Dreissig$^{1,2}$, Dominik Scheuble$^{1}$, Florian Piewak$^{1}$ and Joschka Boedecker$^{2}$
	\thanks{Part of this publication was compiled as part of the research 
		project "KI Delta Learning" (project number: 19A19013A) funded by the Federal 
		Ministry for Economic Affairs and Energy (BMWi) based on a resolution of the 
		German Bundestag.}
	\thanks{$^{1}$Mercedes-Benz AG, $^{2}$University of Freiburg, $^{*}$Primary contact: {\tt\small mariella.dreissig@mercedes-benz.com}}%
}

\makeatletter
\def\endthebibliography{%
	\def\@noitemerr{\@latex@warning{Empty `thebibliography' environment}}%
	\endlist
}
\makeatother

\begin{document}
	
	\maketitle
	\thispagestyle{empty}
	\pagestyle{empty}
	
	\begin{abstract}
		Autonomous vehicles rely on a variety of sensors to gather information about their surrounding. The vehicle's behavior is planned based on the environment perception, making its reliability crucial for safety reasons. The active LiDAR sensor is able to create an accurate 3D representation of a scene, making it a valuable addition for environment perception for autonomous vehicles. Due to light scattering and occlusion, the LiDAR's performance change under adverse weather conditions like fog, snow or rain. This limitation recently fostered a large body of research on approaches to alleviate the decrease in perception performance. In this survey, we gathered, analyzed, and discussed different aspects on dealing with adverse weather conditions in LiDAR-based environment perception. We address topics such as the availability of appropriate data, raw point cloud processing and denoising, robust perception algorithms and sensor fusion to mitigate adverse weather induced shortcomings. We furthermore identify the most pressing gaps in the current literature and pinpoint promising research directions.
	\end{abstract}
	
	\section{Introduction}\label{sec:intro}

The Light Detection and Ranging (LiDAR) sensor recently gained increased attention in the field of autonomous driving \cite{Roriz2021}. It provides sparse but accurate depth information, making it a valuable complement to more well-studied sensors like camera and radar. The LiDAR sensor is an active sensor, meaning it emits light pulses which are reflected by the environment. Afterwards, the sensor captures the reflected light and measures the distance of the environment based on the elapsed time. Additionally to the time, other features can be evaluated, like the amount of light and the elongation of the signal. In most cases, there are mechanical components in combination with multiple laser diodes to create a sparse point cloud of the complete scene \cite{Roriz2021}. There are various different sensors available on the market. 

\begin{figure}[th!]\label{fig:foggy_lidar}
  \centering
  \subfigure{\includegraphics[width=\columnwidth]{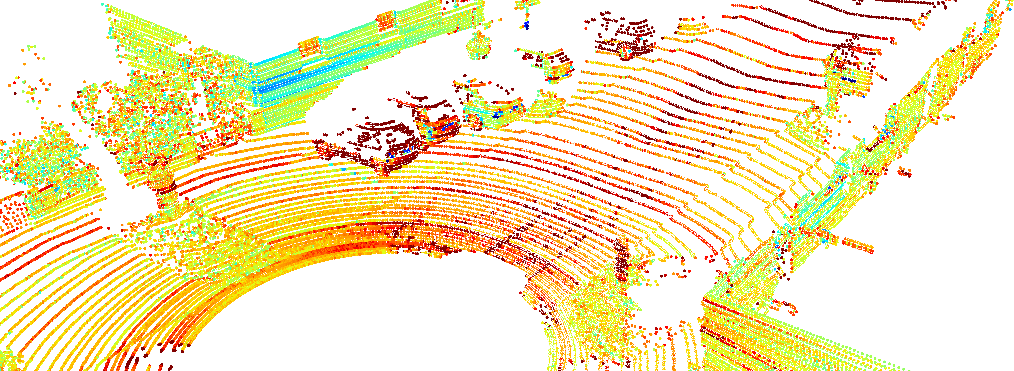}} \\
  \subfigure{\includegraphics[width=\columnwidth]{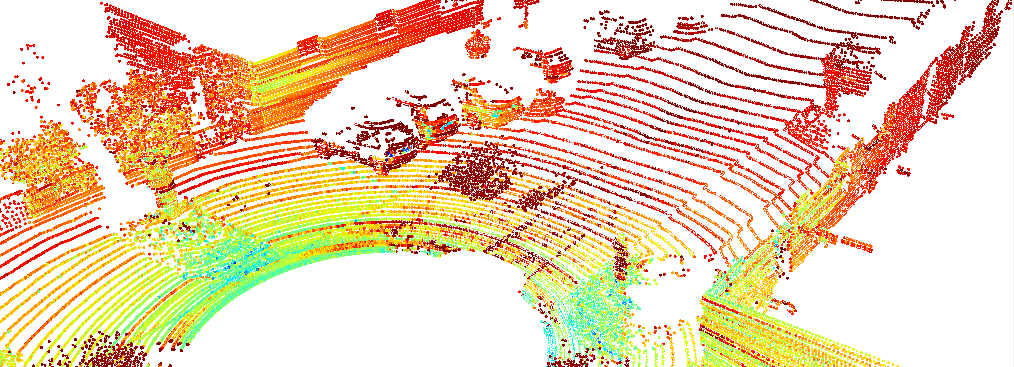}}
  \caption{\textsc{\textbf{LiDAR scans of a street environment under clear weather conditions (top) and with fog (bottom).} The raw data is taken from the KITTI dataset \cite{Geiger2012CVPR} and is augmented with simulated fog and wet road \cite{Hahner2021, Hahner2022}. The color reflects the measured intensity (red = low, blue = high). Noisy measurements can be observed in the proximity of the sensors, which affect the intensity and distribution of the points within the point cloud.}}
\end{figure}

There are different shortcomings for LiDAR sensors under adverse weather conditions. Firstly, sensor freezing or other mechanical complications might occur in freezing temperatures \cite{Kutila2020, Jokela2019}. Internal and structural factors like sensor technology, model and mounting position play a role in the degree of deterioration \cite{Jokela2019}. Additionally, adverse weather affects the intensity values, number of points, and other point cloud characteristics (see Figure \ref{fig:foggy_lidar}) \cite{Montalban2021, Heinzler2022}. In general, when encountering particles in the air due to dust or adverse weather, the emitted light is back scattered or diverted. This results in noisy distance and reflectance measurements in the point cloud, as some laser pulses return too early to the sensor or gets lost in the atmosphere\footnote{It should be noted that e.g. higher rain rates do not necessarily result in lower quality point clouds. The sensor degradation is rather non-linear with respect to different weather characteristics \cite{Montalban2021}.}. 
The noise is especially harmful when applying scene understanding algorithms \cite{Bijelic2018, Heinzler2019}. In this safety-critical usecase it is particularly crucial to maintain a reliably high predictive performance. Thus, there is a need for coping strategies to minimize the LiDAR perception performance degradation under adverse weather conditions, or at least to detect the sensor's limitations in real-world scenarios. 

Most state-of-the-art algorithms rely on deep learning (DL) algorithms, which digest large amounts of data to derive well-generalizing features of the surrounding. While there is a stream of research focusing on unsupervised perception, the majority of recent works require the according labels for the raw data. This includes bounding boxes for object detection and point-wise class labels for semantic segmentation. Manually labeling sparse and additionally noisy point cloud is not only difficult but also expensive and error-prone \cite{Piewak2018}. Thus, the question of how to simulate or augment existing point cloud with weather-specific noise is especially interesting. 

While there is a large body of research on analyzing the performance degradation of LiDAR sensors under adverse weather condition \cite{Montalban2021, Kutila2020, Jokela2019, Bijelic2018, Heinzler2019, YONEDA2019, Abdo2021, Linnhoff2022}, a comprehensive summary on algorithmic coping strategies for an improved perception is missing. Furthermore, surveys on autonomous driving under adverse weather conditions which address weather-induced sensor deterioration \cite{Neumeister2019, Mohammed2020, Zhang2021} do not pinpoint weather-related problems which are unique to the LiDAR sensor. This survey paper summarizes and analyses various approaches on dealing with adverse weather conditions for LiDAR perception. Thereby, we illuminate the topic from three different angles: 
\begin{itemize}
    \item Availability of Data (Section \ref{ch:data}): real-world and synthetic datasets for development of robust LiDAR perception algorithms  
    \item Point Cloud Manipulation (Section \ref{ch:pc_processing}): sensor-specific weather robustness and perception-independent point cloud processing (e.g. weather classification, point cloud denoising)
    \item Robust Perception (Section \ref{ch:perception}): robust perception algorithms which are able to deal with weather-induced noise in the point clouds by e.g. fusing multiple sensors, making adjustments in the training or increasing the general robustness of the perception model
\end{itemize}

 Finally, we provide some concluding remarks about the missing gaps in the current state of the art and also the most promising research directions.
 
	\section{Adverse Weather Data}\label{ch:data}
To train DL models on any kind of perception task, vast amounts of data are required. For supervised approaches, which still dominate the state-of-the-art, those data even have to be labeled through either auto labeling methods or in a manual fashion \cite{Roriz2021}. In either way, obtaining accurately labeled sparse LiDAR data is expensive and cumbersome, and even more impeded when the raw point clouds are corrupted by weather-induced noise. 

Consequently, there is a need for valuable datasets with high quality labels. Generally, there are three options to obtain LiDAR point clouds with weather-characteristic noise patterns: real-world recordings, augmented point clouds and simulated point clouds. The first ones are generated under adverse weather conditions using test cars with appropriate sensor setups. The latter ones require either physical model- or DL-based methods to create parts or the whole point cloud. 

\subsection{Real-world Datasets}
Most of the existing datasets used for LiDAR perception benchmarks are recorded under favorable weather conditions \cite{Geiger2012CVPR, Cordts2016, Sun2019, Caesar2020}. For using developed perception algorithms under real-world situations, the underlying dataset has to reflect all weather conditions. There are some extensive datasets which explicitly include rain, snow and fog besides clear weather conditions. 

\begin{table*}[]\label{tab:datasets}
\centering
\begin{tabular}{@{}lllllllllll@{}}
\toprule
Dataset & \multicolumn{2}{c}{Scenario} & \multicolumn{3}{c}{Weather} & LiDAR & \multicolumn{4}{c}{Ground Truth} \\ \midrule
 & \begin{tabular}[c]{@{}l@{}}open-\\world\end{tabular} & \begin{tabular}[c]{@{}l@{}}weather\\chamber\end{tabular} & rain & fog & snow &  & Motion & \begin{tabular}[c]{@{}l@{}}Object \\ Detection\end{tabular} & \begin{tabular}[c]{@{}l@{}}Semantic \\ Segmentation\end{tabular} & \begin{tabular}[c]{@{}l@{}}Weather \\ Information\end{tabular} \\ \midrule
 
STF \cite{Bijelic2020, Ritter2019}& \checkmark & \checkmark & \checkmark & \checkmark & \checkmark & VLP-32C, HDL64-S3 & - & 4 classes & - & ambient data \\

CADC \cite{Pitropov2020} & \checkmark &  &  &  & \checkmark & VLP-32C & GPS, IMU & 10 classes & - & weather type \\

ADUULM \cite{Pfeuffer2020} & \checkmark &  & \checkmark & \checkmark &  & \begin{tabular}[c]{@{}l@{}}2x VLP-16, \\ 2x VLP-32\end{tabular} & GPS, IMU & - & 12 classes & weather type \\

WADS \cite{Bos2020, Kurup2021} & \checkmark &  &  &  & \checkmark & 2x VLP-16 & GPS, IMU & - & \begin{tabular}[c]{@{}l@{}}22 classes \& 2 \\ snow classes\end{tabular} & point-wise \\ 

LIBRE \cite{Carballo2020} &  & \checkmark & \checkmark & \checkmark &  & \begin{tabular}[c]{@{}l@{}}VLS-128, HDL64-S2,  \\ HDL-32E, VLP-32C, \\ VLP-16, Pandar64, \\ Pandar40P, OS1-64, \\ OS1-16, RS-Lidar32\\ C32-151A, C16-700B\end{tabular} & - & - & - & ambient data \\

\begin{tabular}[c]{@{}l@{}}Oxford RobotCar \\ \cite{Maddern2017, Barnes2020}\end{tabular} & \checkmark &  & \checkmark & \checkmark & \checkmark & 2x HDL-32E & GPS, INS & - & - & - \\

RADIATE \cite{sheeny2020} & \checkmark &  & \checkmark & \checkmark & \checkmark & 2x HDL-32E & GPS & \begin{tabular}[c]{@{}l@{}}2 classes\\ (for radar)\end{tabular} & - & weather type \\

\bottomrule
\end{tabular}
\caption{\textbf{Overview over real-world perception datasets for autonomous driving under adverse weather conditions featuring LiDAR data:} seperated into object detection-, semantic segmenation- and Simultaneous Localization and Mapping (SLAM)-forward datasets. *Weather information explanation: ambient data = exact weather measurements from weather station, weather type = kind / categorical intensity of weather (e.g. fog, (heavy / light) rain / snow, ...), point-wise = point-wise weather label}
\end{table*}

Table \ref{tab:datasets} shows an overview over publicly available datasets for research on LiDAR perception under adverse weather conditions. The datasets are recorded under different conditions and greatly vary in size. Most of them are actually recorded in real-world driving scenarios, while two of them (partially) stem from weather chambers \cite{Bijelic2020, Carballo2020}. Weather chambers have the advantage to fully control the weather conditions and surroundings, i.e. in terms of obstacles. Nevertheless, they do not adequately reflect real-world conditions.

Furthermore, every dataset uses a different sensor setup. \cite{Carballo2020} specifically aims for benchmarking LiDAR manufacturers and models under adverse weather conditions. Besides the LiDAR sensors, all datasets provide RGB camera recordings, some even include radars \cite{Bijelic2020, Maddern2017, sheeny2020}, stereo- \cite{Pfeuffer2020, Maddern2017}, event- \cite{Carballo2020}, gated- \cite{Bijelic2020} or IR-cameras \cite{Bijelic2020, Carballo2020, Bos2020}.

The datasets are designed to tackle different perception and driving tasks for autonomous vehicles. Almost all sensor setups (except \cite{Bijelic2020}) include localization and motion sensors, i.e. GPS/GNSS and IMUs. Thus, they are suitable to develop and test SLAM algorithms. Besides \cite{Barnes2020}, which only provides motion ground truth, all datasets provide either labels for object detection \cite{Bijelic2020, Pitropov2020, Carballo2020, sheeny2020} or point-wise segmentation \cite{Pfeuffer2020, Kurup2021}\footnote{For brevity, the label information are summed up into rough categories.}.

Finally, all datasets include some kind of meta data regarding the weather conditions. This is crucial for the development of almost any kind of perception model under adverse weather conditions. At least for a thorough validation, some knowledge about the intensity and nature of the surrounding weather conditions is crucial. Only one dataset provides point-wise weather label, i.e. falling and accumulated snow along the roadside \cite{Kurup2021}.

The  advantage of datasets consisting of real-world recordings is the high degree of realism. The disadvantages are only partially available (point-wise) labels for the recorded scenes or, in case the data are recorded in a weather chamber, only limited application to way more complex real-world scenarios. The manual point-wise labeling of LiDAR point clouds under adverse weather conditions is especially challenging, since in many cases it is impractical to distinguish clutter or noise from the actual reflection signal. 

\subsection{Weather Augmentation} \label{sub:augmentation}
Augmentation of adverse weather effects into existing datasets provides an effective way of generating vast amounts of data in contrast to tediously collecting and labeling new datasets for different adverse weather effects. Oftentimes, physics-based or empirical augmentation models are used to augment a certain adverse weather effect into clear-weather point clouds, whether they come from real-world drives or from simulations like \cite{Dosovitskiy2017}. This allows to obtain scenes corrupted by weather-specific noise while keeping all interesting edge cases and annotations that are already present in the dataset.

Augmentation methods define a mapping of a clear-weather point to a respective point in adverse weather conditions. To this end, the theoretical LiDAR model in \cite{Rasshofer2011} is often referenced which models the influence of adverse rain, fog and snow. It models the received intensity profile as a linear system by convolving the emitted pulse with the scene response. The scene response models the reflection on solid objects as well as back-scatter and attenuation due to adverse weather. 

A more practical augmentation for fog that can be applied to point clouds directly is introduced in \cite{Bijelic2018}. It is based on the maximum viewing distance which is a function of measured intensity, LiDAR parameters and optical visibility in fog. If the distance of a clear weather point falls below the maximum viewing distance, a random scatter point occurs or the point is lost with a certain probability. This model is adapted to rain by translating visibility parameters and scatter probabilities to rainfall rates \cite{Heinzler2019}. Another rain augmentation model is described in \cite{Goodin2019}. Rain drops are either causing scatter or lost points depending on if the attenuated intensity falls below a noise threshold estimated from the sensor's maximum range. 

Yet, these models ignore the beam divergence of the emitted LiDAR pulse for rain augmentation, which is considered by \cite{Hasirlioglu2018}. Here, the number of intersections of supersampled beams modelling the beam divergence with the spherical rain drops is computed. If the number of intersections exceed a certain threshold, a scatter point is added. The augmentation method in \cite{Teufel2022} extends this approach such that lost points can occur. Furthermore, it is adapted for snow and fog.

Another augmentation for fog, snow and rain is presented in \cite{Kilic2021}. This model operates in the power domain and does not rely e.g. on counting intersections as the previously discussed methods. Additionally, beam divergence is simulated with a computationally more efficient sampling strategy for scatter point distances. In general, the model first compares the attenuated power reflected from solid objects and randomly sampled scatterer with a distance dependent noise threshold. A scatter point is added if the power from scatter points exceeds the one from the solid object. A point is lost if it falls below a distance dependent noise threshold.

Other power domain-driven augmentation methods can be found in \cite{Hahner2021} and \cite{Hahner2022} for fog and snow, respectively. In contrast to \cite{Kilic2021}, they explicitly compute the intensity profile relying on the theoretical formulation from \cite{Rasshofer2011}. Therefore, the different scatterer as well as the solid object contribute to the peak profile. This allows for modelling of occlusions and a more physically accurate augmentation model. Furthermore, \cite{Hahner2022} introduces a wet ground augmentation model that models lost ground points due to the water film on the road. This also allows to estimate the noise floor in a more data-driven way compared to the heuristic one used in \cite{Kilic2021}. The authors of \cite{Liu2022} suggest a physically sound method to estimate both the attenuation and backscattering coefficient to further improve the model proposed in \cite{Hahner2021}.

Aside from physics-based models, empirical models can also be used for augmentation. An empirical augmentation method for spray whirled-up by other vehicles can be found in \cite{LinnhoffSpray2022}. This model is centered around the observation from dedicated experiments that spray is organized into clusters. Another data-driven approach is presented in \cite{Shih2022}, which relies on spray scenes from the Waymo dataset. In \cite{Rivero2021}, a more computationally expensive spray augmentation method is presented that relies on a renderer with a physics engine.

Finally, DL-based methods can be applied to adverse weather augmentation. In \cite{Lee2022}, a Generative Adversarial Networks (GAN)-based approach inspired by image-to-image translation is presented that is able to transform point clouds from sunny to foggy or rainy conditions. They compare their results qualitatively with real foggy and rainy point clouds from a weather chamber.

However, assessing the quality and degree of realism of the augmentation method is challenging. Some authors use weather-chambers or other controlled environments that allow for a comparison with real-world weather effects \cite{Heinzler2019, Carballo2020}. Furthermore, an augmentation method is often considered realistic if it aids the perception performance under real-world adverse weather conditions \cite{Fursa2021}.



	\section{Point Cloud Processing \& Denoising}\label{ch:pc_processing}
In this section, we present approaches on how to deal with adverse weather conditions which are sensor technique- or point cloud-based, i.e. are independent of the actual perception task. Thereby we analyze general sensor-dependent weather robustness and the possibility to estimate the degree of performance degradation depending on the weather conditions. Furthermore, there are streams of research on removing the weather-induced noise from the LiDAR point clouds with both classical denoising methods and DL.  

\subsection{Sensor-related Weather Robustness}
Depending on the technology, the characteristics and the configuration, different LiDAR models are more or less influenced by the weather conditions \cite{Montalban2021, Heinzler2022, Neumeister2019, FILGUEIRA2017}. Due to eye safety restrictions and the suppression of ambient light, two operation wavelengths for LiDAR sensors prevailed: 905nm and 1550nm, with 905nm being the majority of the available sensors. Yet, the 1550nm models appear to have an improved visibility under heavy fog conditions, due to the higher power emit \cite{Kutila2018}. For a thorough discussion on LiDAR technologies under adverser weather conditions, we refer to \cite{Zhang2021}.

Furthermore, the performance Full Waveform LiDAR (FWL) has been investigated under adverse weather conditions \cite{Wallace2020}. FWL measures not only one or two returns but all weaker returns, effectively measuring more noise but also gathering more information about the surrounding. Despite it requires high computational resources, FWL has proven useful to analyse the surrounding medium, which can lay the groundwork for understanding even changing conditions and adjusting dynamically to them.

\subsection{Sensor Degradation Estimation and Weather Classification}
As LiDAR sensors degrade differently under varying weather conditions, estimating the degree of sensor degradation is a first step towards dealing with corrupted LiDAR point clouds. Effords have been made in developing methods to better identify the sensing limits to prevent the propagation of false detections into downstream tasks.

Firstly, some studies on characterizing sensor degradation under various weather conditions \cite{Linnhoff2022, FILGUEIRA2017, Kutila2018} represent a solid basis for sensor calibration under adverse weather conditions and further development, although they are not yet evaluated with regard to their weather classification abilities.

The first work to actually model the influence of rain on the LiDAR sensor is presented in \cite{Goodin2019}. The authors present a mathematical model derived from the LiDAR equation and allow for a performance degradation estimation based on the rain rate and maximum sensing range. 

In subsequent research works, the estimation of the sensor degradation under adverse weather conditions was formulated as an anomaly detection task \cite{Zhang2021Degradation} and a validation task \cite{Delecki2022}. The former employs a DL-based model which aims to learn a latent representation that separates clear from rainy LiDAR scans and thus is able to quantify the degree of the performance decrease. The latter method suggests a reinforcement learning (RL) model to determine failures in an object detection and tracking model. 

While the above-mentioned methods aim to quantify the decrease in the sensor performance itself,  another stream of research focuses on the classification of the surrounding weather conditions (i.e. clear, rain, fog and snow). Satisfying results were achieved with the help of classical machine learning methods (k-Nearest Neighbors and Support Vector Machines) based on hand-crafted features\footnote{The optimal feature set appear to depend on the sensing surface, i.e. the feature set most suitable for classifications based on atmospheric regions might not be the best choice for classifications based on street regions, and vice versa \cite{Rivero2020, Sebastian2021}} from LiDAR point clouds: \cite{Heinzler2019} proposed a feature set to conduct point-wise weather classification, a similar frame-wise approach can be found in \cite{Rivero2020}. 

\cite{Karlsson2022} developed a probabilistic model for frame-wise regressions of the rain rate. With a mixture of experts they accurately infer the rain rate from LiDAR point clouds. 

It should be noted that most of the methods were trained and evaluated on data collected in a weather chamber. While the ability to carefully control the weather conditions allow for high reproducibility, the data usually do not exactly reflect real-world conditions. In order to assess each method's classification abilities, thorough studies on real-world data are necessary \cite{Sebastian2021}.

\subsection{Point Cloud Denoising}\label{sub:denoising}
Weather effects reflect in LiDAR point clouds in terms of specific noise patterns. As described in Section \ref{sec:intro}, they might affect factors like the number of measurements in a point cloud and the maximum sensing range. Instead of augmenting point clouds with weather-specific noise, the point clouds can be denoised by various means in order to reconstruct clear measurements. Additionally to classical filter algorithms, some works on DL-based denoising emerged recently. 

Besides applying perception tasks like object detection on the denoised point clouds, metrics like precision (preserve environmental features) and recall (filter out weather-induced noise) are crucial to evaluate the performance of classical filtering methods. To calculate these metrics, point-wise labels are required which account for weather classes like snow particles \cite{Kurup2021}.

Radius Outlier Removal (ROR) filters out noise based on any point's neighborhood. This becomes problematic for LiDAR measurements of distant objects, as the point cloud becomes naturally sparse. Advanced methods solve this by dynamically adjusting the threshold as a function of the sensing distance (Dynamic Radius Outlier Removal (DROR), \cite{Charron2018, Prio2022}) or taking into account the average distance to each point's neighbors within the point cloud (Statistical Outlier Removal (SOR)). Both methods exhibit high runtimes, making them hardly applicable in autonomous driving. The Fast Cluster Statistical Outlier Removal (FCSOR) \cite{BALTA2018} and the Dynamic Statistical Outlier Removal (DSOR) \cite{Kurup2021} both suggest methods to lower the computational load while still removing weather artifacts from point clouds.

A thorough analysis revealed that weather-induced measurement errors are associated with high density, low intensity, close range and fast decay of points \cite{Wang2022}. Additionally to weather-characteristic neighborhood features, the Low-Intensity Outlier Removal (LIOR) \cite{Park2020} and the Dynamic Distance-Intensity Outlier Removal (DDIOR) \cite{Wang2022} algorithms take measurement intensity into account to remove weather-induced artifacts. The former one utilizes assumptions about the particle size and manually tuned "snow-intensity" threshold, while the latter one aims to unite multiple of the existing filtering ideas into a more sophisticated version. It keeps the computational costs low with the help of a pre-filtering step and achieves compelling results on snowy LiDAR scans.

Denoising methods for roadside LiDARs rely on background models from historical data (which is available for stationary roadside sensors) to identify dynamic points in combination with basic principles used in classical denoising \cite{Wu2020, Sun2022Roadside}. While \cite{Wu2020} filters the weather noise from the actual objects with the help of intensity thresholds (compare \cite{Park2020}), \cite{Sun2022Roadside} filters outliers based on the characteristic local density (compare \cite{Charron2018}). Unfortunately, this is not easily applicable to LiDAR sensors mounted on moving vehicles.

Contrary to classical denoising methods, DL-based denoising of LiDAR point clouds became popular due to the model's abilities to directly understand the underlying structure of weather-induced noise: Firstly, Convolutional Neural Network (CNN)-based models have been used for efficient weather denoising \cite{Heinzler2020, Bergius2022, Yu2022}. The use of temporal data to distinguish further leverages the weather-specific noise removal \cite{Seppaenen2023}, because naturally, the weather noise changes in a higher frequency than the scene background and even the objects within that scene. CNN-based approaches (especially voxel-based) outperform classical denoising methods in terms of noise filtering. Additionally, they have a lower inference time due to faster GPU computations \cite{Bergius2022}. 

Additional to the supervised CNN methods, unsupervised methods like CycleGANs are able to turn noisy point cloud inputs into clear LiDAR scans \cite{Bergius2022}. Yet, they remain noisy in their nature and the resulting point clouds can hardly be validated with respect to their realism \cite{Triess2021}.

	\section{Robust LiDAR Perception}\label{ch:perception}

While there are promising efforts in reducing the domain shift introduced through adverse weather, there are multiple possible approaches on making LiDAR perception models more robust towards adverse weather conditions, independently of the quality and the noise level of the data. There are three streams of work here: utilizing sensor fusion, enhancing training by data augmentation with weather-specific noise, or general approaches on model robustness against domain shifts to compensate performance decrease.

It should be noted that sensor fusion approaches are the only ones tackling multiple perception tasks besides object detection. To the best of our knowledge, there is no literature on other perception tasks like semantic segmentation.  

\subsection{Combating Adverse Weather with Sensor Fusion}
Generally it can be said, that every sensor in an autonomous driving sensor set has its strengths and weaknesses. The most common sensors within such sensor sets are RGB cameras, radars and LiDARs. As discussed in Section \ref{sec:intro}, the LiDAR perception suffers when encountering visible airborne particles like dust, rain, snow or fog. Cameras are more sensitive to strong light incidence and blooming effects. The radar in turn is affected by neither but lacks the capability to detect static objects and finer structures. Thus, it imposes itself to fuse different sensors in order to alleviate their respective shortcomings under different surrounding conditions and facilitate a robust perception. 

Early works on sensor fusion for combating the adverse effect of weather on sensor perception concentrate on the development of robust data association frameworks \cite{Radecki2016, Fritsche2018}. More recent research streams utilize DL-based approaches for robust multi-modal perception and mainly address the question of early vs. late fusion to achieve robustness under adverse weather conditions.

The answer to the question whether to prefer early or late fusion seems to be governed by the choice of the sensors, the data representation, and the expected failure rates. Provided that not all fused sensors are degraded equally and at least one of them is fully functional, late fusion appears to outperform early fusion \cite{Pfeuffer2018, Pfeuffer2019, Rawashdeh2022}. In that case, the model has the ability to treat the sensor streams independently, it is able to rely on the working sensor and ignore the failing one. Contrary, an early fusion of e.g. radar and LiDAR depth maps helps to filter out false detections in order to achieve clean scans \cite{Xie2021} \footnote{Although this work does not explicitly take adverse weather into account, it evaluates the proposed approaches on haze and mist.}. 

The data representation is another factor that partially contributes to answering the question of early vs. late fusion. The Birds Eye View (BEV) of the LiDAR sensor greatly facilitates object detection by improved obejct distinguishability. Thus, any model that has learned to rely on the respective LiDAR features will suffer from a performance loss when the LiDAR data is corrupted \cite{Mirza2021}. Complete sensor failure has successfully been combated by utilizing teacher-student networks \cite{Li2022}.

Ultimately, some sensor fusion approaches rely on combining early and late fusion into one model and exploit concepts like temporal data and region-based fusion \cite{Qian2021} or attention maps \cite{Chaturvedi2022}. Another possibility is the adaptive, entropy-steered fusion proposed in \cite{Bijelic2020}.

Besides the predictive performance, model runtime should also be taken into consideration when developing novel perception approaches \cite{Qian2021}. \cite{Rawashdeh2022} introduced a new metric which incorporates the predictive performance for drivable space segmentation with the inference runtime. Interestingly, the LiDAR-only model scored best on that metric.

Undoubtedly, it is convenient to compensate sensor failure under adverse weather conditions with unaffected sensors\footnote{\cite{Elmassik2022} proposes optimized sensor setups}. Yet, by striving for improving the LiDAR-only perception under adverse weather conditions, safety-critical applications like autonomous driving can become even more reliable.

\subsection{Enhancing Training with Data Augmentation}
While data augmentation is widely used in DL training strategies, it is the creation of specific weather noise which is particularly challenging. Section \ref{sub:augmentation} presented a variety of methods to generate weather-specific noise in LiDAR point clouds. Utilizing data augmentation during the training of a perception model is the diametrical method of point cloud denoising, which has been discussed in \ref{sub:denoising}. Instead of removing the weather-induced noise, the aim is to make the model accustomed to that exact noise. It has been demonstrated that weather augmentation is more effective than denoising in terms of robustness, which gives valuable hints on which research direction should be emphasized in the future \cite{Hahner2022}. 

Generally, several works demonstrate the benefits of such data augmentation at training time by evaluating them on the task 3D object detection \cite{Hahner2021, Hahner2022, Kilic2021}. 

Many works address the subject of choosing the best feature extractor for robust LiDAR perception under adverse weather conditions. Point-based and voxelizing methods appear to be less affected by the augmented weather effects \cite{Hahner2021, Hahner2022, Kilic2021}, at least for object detection, hinting that some robustness can be achieved by carefully choosing the perception model. Also, there seems to be an interaction between the model architecture and the kind of point cloud corruption due to adverse weather. The wet ground extension presented in \cite{Hahner2022} only aids some models, indicating that the detection problems caused by ray scattering are more or less grave, depending on the model architecture.  

Furthermore, the size and shape of objects seem to play a role in the degree of any detection model's performance degradation \cite{Hahner2021, Hahner2022, Vattem2022}. That means, smaller and underrepresented classes (like cyclist in the STF dataset) suffer more from the weather augmentation than well-represented classes, like car and pedestrian. Thus, the number of annotated objects in the (clear) training set is a good indicator on the object detection performance even under adverse weather conditions. This indicates that not only training with weather augmentation aids the detection performance under clear weather conditions \cite{Hahner2022}, interestingly, it also appears to work inversely \cite{Vattem2022}.

\subsection{Robust Perception Algorithms}
While fusion methods with complementary sensors alleviate the weather-induced performance degradation of each single sensor, they only act as a workaround for the actual problem at hand. Changes in the weather conditions can be seen as a special case of domain shift \cite{Sun2022}, thus approaches developed to bridge domain gaps might be applied to the weather-to-weather (e.g. clear-to-rain/fog/snow) domain shift \footnote{\cite{triess2021iv} gives a comprehensive overview over the current state of the art domain adaptation methods, but they mainly tackle problems related to different sensor resolutions or the available data and their labels.}. Since there are no extensive datasets adressing the weather-to-weather domain shift only, it can be evaluated as part of the dataset-to-dataset domain shift. Thus, two works on developing robust LiDAR pereption algorithms indirectly evaluate the performance under adverse weather conditions. While the works provide interesting insights into the problem at hand, it should be noted, that since the domain gap was not limited to the shift between weather conditions, other factors like sensor resolution and label strategy might overshadow the weather-induced gap. Thus, in the evaluation it is unclear which elements of the model attribute to the shift in the weather condition itself, since the dataset-to-dataset shift is very strong.

\cite{Caine2021} employ a teacher-student-setup for object detection where the teacher is trained on Waymo Open (sunny) to generate labels for part Waymo Open, part Kirkland (rainy), student is trained on all label and applied to Kirkland. Interestingly, the students appeared to generalize better to the target domain, indicating that they were able to cope with the adverse weather. The authors of \cite{Lin2022} proposed to robust object detection pipeline including attention mechanisms and global context-aware feature extraction which allows the model to ignore weather-induced noise and at the same time, understand a whole scene. While their methods fail to perform well on two domains simultaneously (KITTI, sunny \& CADC, rainy), a joint training based on a maximum discrepancy loss yields promising results and shows high performances on both source and target domain. 

\cite{Eskandar2022} focuses on alleviating weather-induced sensor degradation for both RGB camera and LiDAR. Although they utilize sensor fusion (derived from the entropy fusion presented in \cite{Bijelic2020}) as well as data augmentation for both sensors, their work strongly contributes towards exploiting a set of methods to bridge the gap to multiple unknown target domains for object detection. They achieve this by introducing domain discriminators and domain alignment by self-supervised learning through a pre-training strategy. Their results show that their multi-modal, multi-target domain adaptation method is able to generalize well to e.g. fog scenarios. 

	\section{Discussion and Conclusion}
In this survey paper we outlined current research directions in LiDAR-based environment perception for autonomous driving in adverse weather conditions. We thoroughly analyzed and discussed the availability of training data for deep learning algorithms, perception-independent point cloud processing techniques for detecting weather conditions and denoising the LiDAR scans, and finally, current state-of-the-art approaches on robust LiDAR perception. In the following, we will summarize most promising research directions and identify remaining gaps.

\textit{Adverse Weather Data} (Section \ref{ch:data}): There are several autonomous driving datasets which include LiDAR sensors and simultaneously cover adverse weather conditions. Most of them provide object labels, but only one has point-wise class labels. There clearly is a need for appropriate real-world datasets to train and validate the growing amount of deep learning based LiDAR perception algorithms. Some works resort to weather-specific data augmentation to simulate adverse weather effects, yet, a method to evaluate the realism of the generated augmentations is missing.

\textit{Point Cloud Processing \& Denoising} (Section \ref{ch:pc_processing}): Distinct LiDAR technologies react differently to adverse weather conditions. While thorough studies on sensor degradation under adverse weather conditions exist, a systematic analysis of the impact on perception algorithms is missing. Here, approaches on sensor degradation estimation will be useful. Furtheremore, there is ongoing research on cloud denoising, but existing statistical methods have been proven less efficient than utilizing weather augmentation during training. Modern methods like CNN- or GAN-based approaches might bridge that gap.

\textit{Robust LiDAR Perception} (Section \ref{ch:perception}): A large body of research focuses on alleviating sensor degradation with the help of sensor fusion. While this yields compelling results, improving the LiDAR-only perception under adverse weather conditions should not be neglected. Sophisticated domain adaptation approaches (like anomaly detection or uncertainty modeling) might be useful to address that matter. Viewing the presence of weather-induced noise in LiDAR point clouds from different perspectives might unlock novel streams of research on bridging the domain gap introduced by adverse weather conditions. Investigating the quality of that domain gap would give hints on the potential of general domain adaptation approaches.

	\addtolength{\textheight}{-0.5cm}   

	\bibliographystyle{bibliography/IEEEtran}
	\bibliography{bibliography/refs}
	
\end{document}